\documentclass[runningheads]{llncs}
\usepackage{graphicx}
\usepackage{booktabs}
\usepackage{multirow}
\usepackage{tikz}
\usepackage{comment}
\usepackage{amsmath,amssymb} 
\usepackage{bbm}
\usepackage{color}
\usepackage{epsfig}
\usepackage{epstopdf}
\usepackage{diagbox}

\usepackage[colorlinks,linkcolor=red]{hyperref}

\begin{document}
\pagestyle{headings}
\mainmatter
\def\ECCVSubNumber{ }  

\title{Mid-level Representation Enhancement and Graph Embedded Uncertainty Suppressing for Facial Expression Recognition} 


\titlerunning{4th ABAW Competition: LSD Challenge}
%
\author{Jie Lei\inst{1} \and
Zhao Liu\inst{2} \and
Zeyu Zou\inst{2} \and
Tong Li\inst{2} \and
Xu Juan\inst{2} \and
Shuaiwei Wang\inst{1} \and
Guoyu Yang\inst{1} \and
Zunlei Feng\inst{3}}
\authorrunning{J. Lei et al.}
%
\institute{Zhejiang University of Technology, Hangzhou, 310023, P.R. China
\email{\{jasonlei,swwang,gyyang\}@zjut.edu.cn}\\ \and
Ping An Life Insurance Of China, Ltd, Shanghai, 200120, P.R. China
\email{\{liuzhao556,zouzeyu313,litong300,xujuan635\}@pingan.com.cn}\\ \and
Zhejiang University, Hangzhou, 310027, P.R. China\\
\email{zunleifeng@zju.edu.cn}}
\maketitle

\begin{abstract}
Facial expression is an essential factor in conveying human emotional states and intentions. Although remarkable advancement has been made in facial expression recognition (FER) task, challenges due to large variations of expression patterns and unavoidable data uncertainties still remain. In this paper, we propose mid-level representation enhancement (MRE) and graph embedded uncertainty suppressing (GUS) addressing these issues. On one hand, MRE is introduced to avoid expression representation learning being dominated by a limited number of highly discriminative patterns. On the other hand, GUS is introduced to suppress the feature ambiguity in the representation space. The proposed method not only has stronger generalization capability to handle different variations of expression patterns but also more robustness to capture expression representations. Experimental evaluation on Aff-Wild2 have verified the effectiveness of the proposed method. We achieved 2nd place in the Learning from Synthetic Data (LSD) Challenge of the 4th Competition on Affective Behavior Analysis in-the-wild (ABAW). The code will be released at \url{https://github.com/CruiseYuGH/GUS}.

\end{abstract}

\section{Introduction}
Facial expression recognition (FER) is a highly focused field in computer vision for its potential use in sociable robots, medical treatment, driver fatigue surveillance, and many other human-computer interaction systems. Typically, the FER task is to classify an input facial image into one of the following six basic categories: \emph{anger} (AN), \emph{disgust} (DI), \emph{fear} (FE), \emph{happiness} (HA), \emph{sadness} (SA), and \emph{surprise} (SU).

In the field of facial expression recognition (FER), various FER systems have been explored to encode expression information from facial representations. Several studies have proposed well-designed auxiliary modules to enhance the foundation architecture of deep models~\cite{2016HoloNet,2018Patch,2017Learning,2021Identity,2020Facial}.
Yao \emph{et al.}~\cite{2016HoloNet} proposed HoloNet with three critical considerations in the network design. Zhou \emph{et al.}~\cite{2020Facial} introduced an spatial-temporal facial graph to encode the information of facial expressions.
Another area focuses on facial expression data for robust recognition~\cite{2020Suppressing,2018Facial2,2016Peak,2018Facial,2017Contrastive}.
In~\cite{2018Facial2}, the authors proposed an end-to-end trainable LTNet to discover the latent truths with the additional annotations from different datasets. Zhao \emph{et al.}~\cite{2016Peak} presented a novel peak-piloted deep network (PPDN) that used the peak expression (easy sample) to supervise the non-peak expression (hard sample) of the same type and from the same subject.

Despite the promising results, prior approaches tend to learn a limited number of highly discriminative patterns and suffer from unavoidable expression data uncertainties. To address these issues, we propose to mix the mid-level representation with samples from other expressions in the learning process, thus suppressing the over-activation to some partial expression patterns and enriching the pattern combination to enhance classifier generalization. In addition, we introduce graph embedded uncertainty suppressing to reduce the feature ambiguity in the representation space.

The competition on Affective Behavior Analysis in-the-wild (ABAW) has been successfully held three times, in conjunction with IEEE FG 2020, ICCV 2021, and CVPR 2022. The goal of our work is to study facial expression recognition based on Aff-Wild2~\cite{kollias2022abaw,kollias2021distribution,kollias2021affect,kollias2020deep,kollias2020va,kollias2019expression,kollias2019deep,kollias2018photorealistic,zafeiriou2017aff,kollias2017recognition,kollias2022abaw4} in the 4th competition. We adopted the proposed mid-level representation enhancement (MRE) and graph embedded uncertainty suppressing (GUS) for this FER task. Experiments on Aff-Wild2 show the effectiveness of the proposed method.

\section{Method}
\subsection{Overview}
\begin{figure}
\centering
\begin{center}
\centerline{\epsfig{file=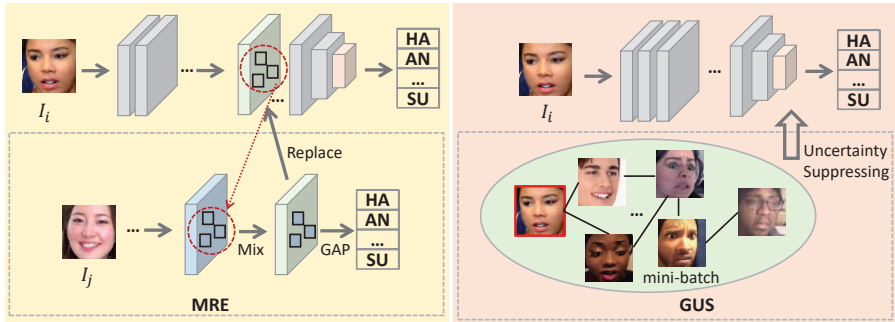,width=12cm}}
\end{center}
\caption{Framework of the proposed MRE and GUS.}
\label{fig:framework}
\end{figure}

The overall architecture of our proposed method is illustrated in Fig.~\ref{fig:framework}. It mainly consists of two parts, \emph{e.g.}, mid-level representation enhancement and graph embedded uncertainty suppressing. In order to further enhance the robustness of our framework, we adopt the model ensemble strategy by combining DMUE~\cite{She2021DMUE} with MRE and GUS.

\subsection{Mid-level Representation Enhancement}
The mid-level representation can support robust high-level representation to improve accuracy~\cite{2021SPS}. However, the learned mid-level representation tends to be dominated by a limited number of highly discriminative partial patterns, making false predictions when two expressions share significant similarities in specific facial parts that are regular in one category.

For facial image $I_i$, based on the hierarchical structure of deep neural networks, we can take the expression representations of mid-level and high-level according to the layer depth, denoted as $R^m_i$ and $R^h_i$, respectively.

We mix the mid-level representation with a sample from other expressions to form an new variant. The main idea is to inject partial features as noise data to improve the generalization capability. Given a facial image $I_i$ and a randomly selected sample $I_j$ with a different expression, the mixing operation can be denoted as:
\begin{equation}
\label{eqn:eqn1}
\tilde{R_i^m}(p)=\mathbbm{1}[p \notin \mathcal{N}] R_i^m(p)+\mathbbm{1}[p \in \mathcal{N}] R_j^m(p),
\end{equation}
where $\mathbbm{1}(\cdot)$ is an indicator, $\mathcal{N}$ is the uniform sampled positions for where the noises are injected, $\tilde{R_i^m}(p)$ indicates the value of the variant of $R_i^m$ at position $p$. This strategy provides a way to allow gradients to suppress the overconfident partial patterns.

Following the procedure, the variant $\tilde{R_i^m}$ is obtained, which helps prevent the network from over-focusing on a small number of discriminative partial patterns. To amplify this property, we introduce an expression classification module based on $\tilde{R_i^m}$ as an additional branch via global average pooling (GAP) in the network. In this way, the mid-level expression representation can be enhanced by directly supervised with the expression label. Thus, the total loss for the FER task is defined as:
\begin{equation}
\label{eqn:eqn2}
\mathcal{L}_{MRE}=\sum_{i} (\ell(R^h_i, y_i) + \lambda \ell(\tilde{R_i^m}, y_i)),
\end{equation}
where $y_i$ is the expression label for $I$, the first term denotes the original training loss for expression classification towards the high-level expression representation, the second term denotes the expression classification loss using the variant mid-level representation on the branch, and $\lambda$ is a balanced parameter.

\subsection{Graph Embedded Uncertainty Suppressing}
For uncertainty suppressing, we use Graph Convolutional Networks (GCN)~\cite{Kipf:2016tc,AMGCN} to estabilish semantic relations between the repesentations and reduce the covariate shifting. The inputs are representation maps $\hat{F} = \{\hat{f_1},..., \hat{f_N}\}$ extracted by backbone feature extractor from the original images. To begin with, we use cosine similarity coefficient to calculated the similarity between different representations as:

\begin{equation}
\label{eqn:eqn1}
cossim(f_i, f_j) = \frac{f_i*f_j}{||f_i||f_j||},
\end{equation}

then, we employ GCN as:

\begin{equation}
F^{l+1} = \widetilde{D}^{-\frac{1}{2}}\widetilde{A}\widetilde{D}^{-\frac{1}{2}}F^{l}W^{l},
\end{equation}

where $\widetilde{A} = A + I$ is the sum of undirected graph A obtained by Eqn.~(\ref{eqn:eqn1}) and the identity matrix, $\widetilde{D}$ is the diagonal matrix from A, which is $\widetilde{D}_{i,i} = \sum\limits_{j}A_{i,j}$. $F^{l}$ and $F^{l+1}$ are the corresponding input and output representations on the $l_{th}$ level, and $W^{l}$ are the trainable parameters on this level.

\subsection{Mixed Loss Functions}
In the training process, we have conducted multiple combinations of different loss functions, finally we use the combination of cross entropy loss $L_{ce}$, focal loss $L_{fl}$~\cite{2017FocalLoss} and sparse regularization loss $L_{fl}$~\cite{2021SRLoss}. Without loss of generality, the predicted outputs of our model is $\hat{Y} = \{\hat{y_1},..., \hat{y_6}\}$ the ground truth label is $Y = \{y_1,...y_6\}$, our final loss function can be formulated as:
\begin{equation}
L_{fer} = \omega_1L_{ce} + \omega_2L_{fl} + \omega_3L_{sr},
\end{equation}
where $\omega_1$, $\omega_2$ and $\omega_3$ are balanced parameters.

\subsection{Predictions Merging and Correcting}
We employ a predictions merging strategy to further enhance the robustness of the whole framework. For the features extracted from different sub-networks, we use softmax function to calculate the confidential scores of each expression category, then the confidential scores of each class are summed up as:

\begin{equation}
R_{final} = \omega^r_gR_{GUS} + \omega^r_mR_{MRE} + \omega^r_dR_{DMUE},
\end{equation}

where weights $\omega^r_g$, $\omega^r_m$ and $\omega^r_d$ are used to balance the imfluence of different sub-networks in the final predictions. During the training process we found GUS achieved better results than DMUE and MRE in all the expression categories except \emph{fear}. As a consequence, we design two different weights settings, denoted as S1 and S2. S1: $\omega^r_g = 0.6$, $\omega^r_m = 0.2$ and $\omega^r_d = 0.2$, convicing more on the predictions of GUS. S2: $\omega^r_g = 0.4$, $\omega^r_m = 0.3$ and $\omega^r_d = 0.3$, relying more on the predictions of MRE, DMUE. Finally the \emph{argmax} function is employed to predict the final outputs.

In addition, we design a predictions correcting strategy for the test set. We adopt a Imagenet pre-trained VGG-19 model to extract features from the test images, and calculate the cosine similarity between different features. The images with similarity higher than 0.93 are classified to the same subsets. During the inference stage, a voting process is carried out in each subset with more than 2 images: If more than two-thirds of the images are predicted as the same category, the predicted results of the remaining images are updated with this category.

\section{Experiments}
\subsection{Dataset}
The Learning from Synthetic Data (LSD) Challenge is part of the 4th Workshop and Competition on Affective Behavior Analysis in-the-wild (ABAW). For this Challenge, some specific frames-images from the Aff-Wild2 database have been selected for expression manipulation. In total, it contains 277,251 synthetic images annotations in terms of the six basic facial expressions. In addition, 4,670 and 106,121 images from Aff-Wild2 are selected as the validation and test sets. The synthetic training data have been generated from subjects of the validation set, but not of the test set.

\subsection{Settings}
We only use the officially provided synthetic images to train the models and the validation images for finetuning. All the images are resized to 224 $\times$ 224 in these processes. We use random crop, Gaussian blur and random flip as the data augmentation strategies. We also adopt over-sampling strategy over the classes with less images to reduce data imbalance. For training the MRE branch, we use ResNet-50 as the backbone network. As for the DMUE~\cite{She2021DMUE} and GUS branches, we adopt ResNet-18 as the backbone network. All the backbones are pre-trained on ImageNet. We use SGD with a learning rate of 0.001 during the optimization in MRE and DMUE, and a learning rate of 0.0001 in GUS.

\subsection{Results}
Table~\ref{tab:aFER} shows the results of the proposed method on the official validation dataset. By employing DMUE~\cite{She2021DMUE}, MRE and GUS, we achieve 0.28, 0.14 and 0.32 increase on mean F1-score compared with the baseline model (Resnet-50), respectively. By applying the model ensemble strategy, we can achieve another 0.03 increase. Meanwhile, the predictions correcting stratagy contributes another 0.02 increase. Finally, our proposed framework scored 0.8482 on the validation dataset.

Table~\ref{tab:Final} shows the results on the test set of 4th ABAW Learning From Synthetic Data (LSD) Challenge. The five submissions are obtained as follows: (1) Single GUS module; (2) Predictions merging with ensemble weights S1; (3) Predictions merging with ensemble weights S2; (4) Predictions merging with ensemble weights S1 and predictions correcting; (5) Predictions merging with ensemble weights S2 and predictions correcting. As shown in the table, the 4th submission owns the highest performance of 36.51 in F1-score, and we achieve 2nd place in the final challenge leaderboard.

\begin{table*}
\scriptsize
\begin{center}
\begin{tabular}{|l|c|c|c|c|c|c|c|}
\hline
\diagbox{Method}{Class} &ANGER &DISGUSS &FEAR &HAPPINESS &SADNESS &SURPRISE &MEAN \\
\hline\hline
Baseline &0.4835&0.5112&0.3860&0.5022&0.4600&0.5731&0.4865\\
DMUE~\cite{She2021DMUE}&0.8235&0.8000&0.6768&0.7438&0.7541&0.7929&0.7642\\
MRE&0.7797&0.4737&0.4521&0.7061&0.6768&0.6809&0.6282\\
GUS&0.8846&0.8814&0.5578&0.8616&0.8399&0.8315&0.8095\\ \hline
Ensembled&0.8846&0.8814&0.6909&0.8616&0.8399&0.8558&0.8357\\
Corrected&0.8846&0.8966&0.7143&0.8584&0.8674&0.8676&0.8482\\ \hline
\end{tabular}
\end{center}
\caption{Performances of different models and strategies on the official validation dataset, where all the results are reported in F1-Score.}
\label{tab:aFER}
\end{table*}

\begin{table*}
\begin{center}
\begin{tabular}{|c|c|}
\hline
Submission & Performance Metric \\
\hline\hline
1 & 31.54 \\
2 & 36.13 \\
3 & 36.33 \\
4 & \textbf{36.51} \\
5 & 36.35 \\ \hline
\end{tabular}
\end{center}
\caption{Performance of 5 submissions on the test set of 4th ABAW Learning From Synthetic Data Challenge (LSD).}
\label{tab:Final}
\end{table*}

\section{Conclusion}
In this paper, we have proposed the mid-level representation enhancement (MRE) and graph embedded uncertainty suppressing (GUS) for facial expression recognition task, aiming at addressing the problem of large variations of expression patterns and unavoidable data uncertainties. Experiments on Aff-Wild2 have verified the effectiveness of the proposed method.

\clearpage
%
%
\bibliographystyle{splncs04}
\bibliography{abaw2022lsd}
\end{document}